\documentclass[runningheads]{llncs}

 \usepackage[accsupp]{axessibility} 
\usepackage{eccv}

\usepackage{eccvabbrv}

\usepackage{graphicx}
\usepackage{booktabs}

\usepackage{hyperref}

\usepackage{orcidlink}

\usepackage{amsmath,graphicx}
\usepackage{pifont}
\newcommand{\cmark}{$\checkmark$}%
\newcommand{\xmark}{}%
\usepackage{booktabs}
\usepackage[acronym,nomain,nonumberlist]{glossaries}
\glsdisablehyper
\newcommand{\norm}[1]{\left\lVert#1\right\rVert}
\newcommand{\abs}[1]{\left\lvert#1\right\rvert}

\newacronym{ct}{CT}{Computed Tomography}
\newacronym{mri}{MRI}{Magnetic Resonance Imaging}
\newacronym{log}{LoG}{Laplacian of Gaussian}
\newacronym{3D}{3D}{3-dimensional}
\newacronym{2D}{2D}{2-dimensional}
\newacronym{MLP}{MLP}{Multilayer Perceptron }
\newacronym{PCA}{PCA}{Principal Component Analysis}
\newacronym{MAD}{MAD}{Median Absolute Deviation}
\newacronym{CNN}{CNN}{Convolutional Neural Network}
\newacronym{IoU}{IoU}{Intersection over Union}

\begin{document}

% ---------------------------------------------------------------
% TODO REVIEW: Replace with your title
\title{Deep Unsupervised Segmentation of Log Point Clouds} 

% TODO REVIEW: If the paper title is too long for the running head, you can set
% an abbreviated paper title here. If not, comment out.
% \titlerunning{Abbreviated paper title}

% TODO FINAL: Replace with your author list. 
% Include the authors' OCRID for the camera-ready version, if at all possible.
\author{Fedor Zolotarev\inst{1}\orcidlink{0009-0008-1978-5764} \and
Tuomas Eerola\inst{1}\orcidlink{0000-0003-1352-0999} \and
Tomi Kauppi\inst{2}}

% TODO FINAL: Replace with an abbreviated list of authors.
\authorrunning{F.~Zolotarev et al.}
% First names are abbreviated in the running head.
% If there are more than two authors, 'et al.' is used.

% TODO FINAL: Replace with your institution list.
\institute{Lappeenranta-Lahti University of Technology LUT, Lappeenranta, Finland \\
\email{firstname.lastname@lut.fi} \and
Finnos Oy, Lappeenranta, Finland \\
\email{firstname.lastname@finnos.fi}}

\maketitle

\begin{abstract}
  In sawmills, it is essential to accurately measure the raw material, i.e. wooden logs, to optimise the sawing process. Earlier studies have shown that accurate predictions of the inner structure of the logs can be obtained using just surface point clouds produced by a laser scanner. This provides a cost-efficient and fast alternative to the X-ray CT-based measurement devices. The essential steps in analysing log point clouds is segmentation, as it forms the basis for finding the fine surface details that provide the cues about the inner structure of the log. We propose a novel Point Transformer-based point cloud segmentation technique that learns to find the points belonging to the log surface in unsupervised manner. This is obtained using a loss function that utilises the geometrical properties of a cylinder while taking into account the shape variation common in timber logs. We demonstrate the accuracy of the method on wooden logs, but the approach could be utilised also on other cylindrical objects.
\end{abstract}

\section{Introduction}
\label{sec:intro}

Improving the sawmill processes is an ongoing effort that depends on new technologies and methods for measuring the raw material, i.e. wooden logs. Improving board production efficiency not only reduces wood wastage but also contributes to a decline in excess logging, fostering sustainable forestry practices. Therefore, the process optimisation plays a pivotal role in promoting rational forest resource management and enhancing the revenues of sawmill companies. 

One of the key steps to optimise is the determination of the best sawing strategy for each log, maximising the quality and value of the wood products. The optimisation problem can be formulated as finding the sawing pattern (dimensions and arrangement of the planks and boards) and the sawing angle that maximise the profit. Various studies have demonstrated the benefits of optimal sawing parameters on profitability and yield~\cite{rais2017use,stangle2015potentially}. A very useful tool for sawing optimisation is the so-called virtual sawing, that is predicting the locations of defects based on the log measurements. This, together with feedback loops, enables using timber tracing methods (see e.g.~\cite{zolotarev2019timber}), providing novel possibilities for sawmill process optimisation. To ensure accurate virtual sawing, obtaining precise log measurements as well as general (sawmill independent) log data processing techniques is essential. A wide variety of different sawmill production lines with vastly different measurement systems render supervised approaches for log data processing infeasible as collecting and annotating enough sawmill specific training data is simply not possible. This calls for unsupervised approaches to obtain representative and meaningful log representations despite the differences in sawmills.

\begin{figure}[t!]
  \centering
  \begin{subfigure}{0.49\linewidth}
    \includegraphics[width=1\linewidth]{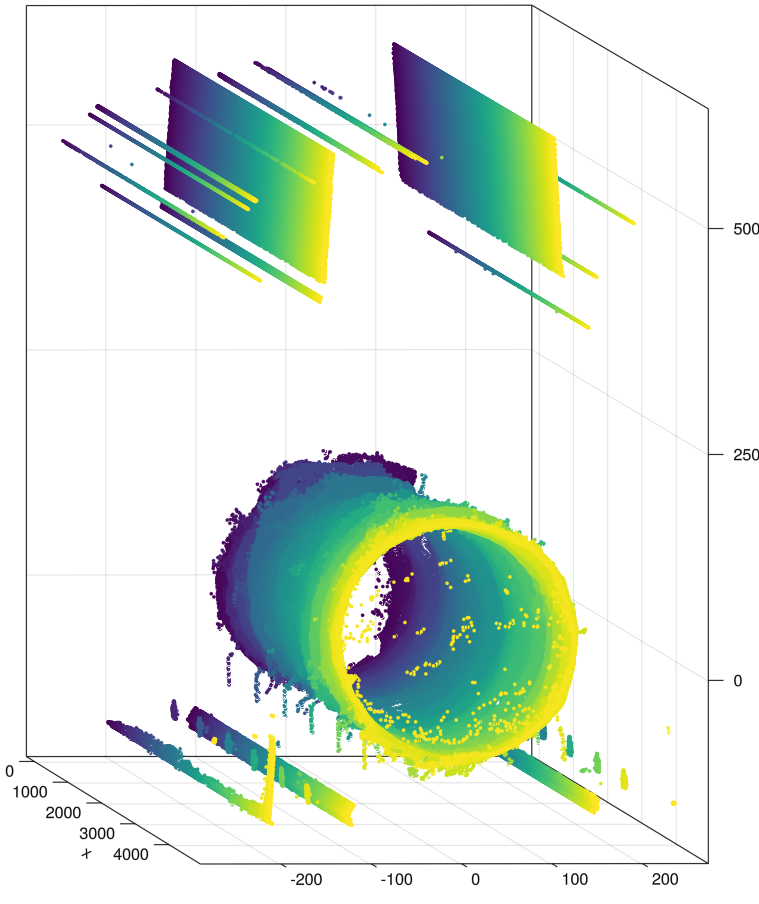}
    \caption{Raw point cloud of a log.}
  \end{subfigure}
  \hfill
  \begin{subfigure}{0.49\linewidth}
    \includegraphics[width=1\linewidth]{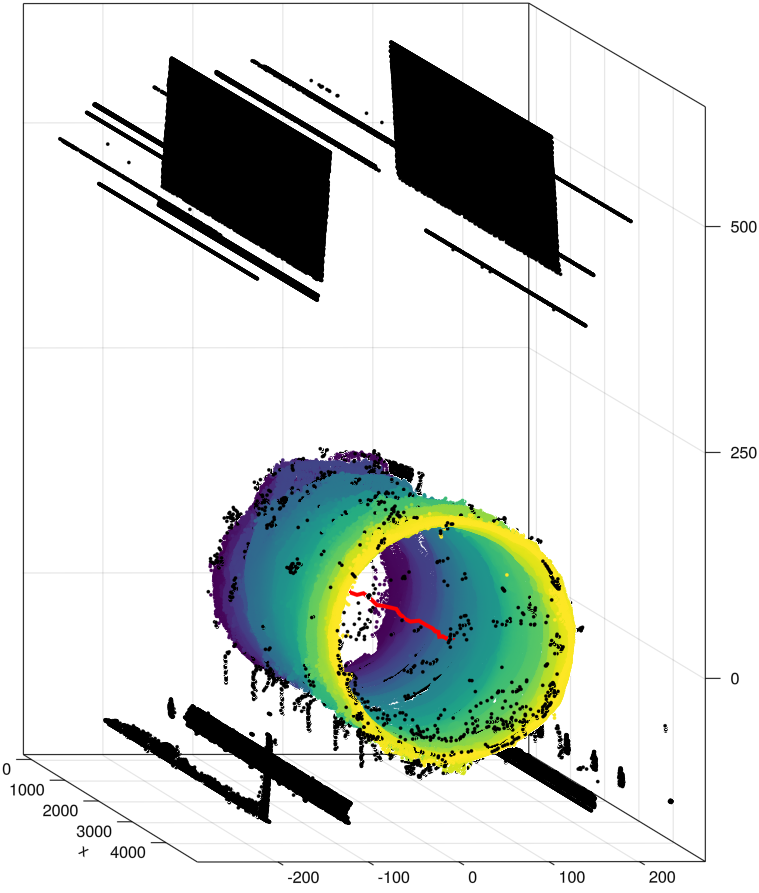}
    \caption{Segmented log.}
  \end{subfigure}
  \caption{Example of a correctly segmented log.}
  \label{fig:example}
\end{figure}

In the context of sawing optimisation, one approach to acquire essential information is through the use of advanced sensors. While \gls{ct} or \gls{mri} scanners are able to extract detailed information about the internal structure of the log, they are often too expensive or too slow to be used in the real industrial setting. On the other hand, laser range scanners are much faster and cost-effective, even though they are limited to providing the information about the external features of the log. However, it has been shown in~\cite{zolotarev2019timber,zolotarev2020modelling} that even surface level information can be used to solve the problems of virtual sawing and timber tracing. One of the drawbacks of the existing solutions is the lack of computation speed while processing point clouds, which defeats the purpose of using high-speed sensors in a real-time setting. By replacing some steps of the pipeline with an end-to-end neural network it is possible to greatly speed up the computation. The main challenge is the lack of ground truth labels, which is exceedingly difficult to collect. This is why an unsupervised method would present an ideal solution for addressing this issue.

This study presents a novel approach for  unsupervised segmentation of a log point cloud. An example of input data and desired output is presented in \cref{fig:example}. This method can be, for example, used as a replacement step for the methods proposed in~\cite{zolotarev2019timber,zolotarev2020modelling} for timber tracing and virtual sawing, respectively. Both methods begin with segmentation aiming to provide accurate reconstruction of the surface for further detection of subtle defects. 
The problem with previously proposed solutions is that they are relatively slow to compute and have a lot of parameters that require manual tuning. Moreover, the manual tuning needs to be done again if the measurement setup changes or the method is applied in a new sawmill environment. Those issues can be addressed by using a neural network-based solution which utilises unsupervised learning and geometrical properties of a log shape. The main contribution of this paper is a new unsupervised loss for the segmentation of log point clouds. We demonstrate the accuracy of the proposed method on three challenging log point cloud datasets collected from different sawmills. It also bears mention that cylindrical shapes are not exclusive to logs and are commonly encountered in both industrial setting and nature. While the proposed method is developed for timber logs, it could also be applied to other domains.

\section{Related work}

\subsection{Segmentation of Log Point Clouds}
Various methods have been proposed to segment both point clouds of timber logs~\cite{thomas2007robust,thomas2009modeling,zolotarev2019timber, zolotarev2020modelling} and terrestrial laser scans of trees~\cite{pfeifer2004automatic,kretschmer2013new}. In the context of tree trunks and logs, the point cloud segmentation refers to finding the points that correspond to the log surface and ignoring the outliers such as noise and artefacts from the environment.

In~\cite{pfeifer2004automatic}, a method to fit and track cylinders to reconstruct a tree from terrestrial scans was proposed. The method was further developed in~\cite{kretschmer2013new} by introducing the conversion to a cylindrical coordinate system, which allows the representation of the trunk or log surface as heightmaps where possible defects can be detected.

A series of works by Thomas~\etal~\cite{thomas2006automated,thomas2007robust,thomas2009modeling,thomas2013raysaw} applied similar techniques to timber logs. The general idea is to fit circles to each cross section of a log. Distances of points to the circles are calculated to obtain a 2D surface map where coordinate axes correspond to the angle around the log and the position along the length of the log. 

In~\cite{zolotarev2019timber, zolotarev2020modelling, batrakhanov2021virtual}, the presence of outliers was addressed by clustering the point clouds for each cross-section using the DBSCAN method~\cite{ester1996density}. This is based on the assumption that the densest cluster corresponds to the log surface. Further, noise filtering is achieved by fitting a circle to the detected cluster points using least squares fitting and discarding the outliers based on a predefined threshold for the distance to the fitted circle.

All the above methods make the assumption of a strictly cylindrical shape or circular cross-sections. This is often not the case and for logs with more irregular shapes these methods might fail to correctly segment the log.

\subsection{Point Clouds of Cylindrical Objects}

Detection of cylindrical objects has many other industrial applications. The most common use case is modelling of pipes in industrial plants~\cite{young2019fast, patil2017adaptive, lee2013skeleton, youngdoo2020automatic, agapaki2020cloi, oh2021automated}. In industrial setting, the objects to be segmented are man-made and therefore have strictly defined models, allowing to use approaches based on Hough-transform~\cite{patil2017adaptive, romanengo2023recognizing} or RANSAC~\cite{young2019fast, oh2021automated}. Before fitting the pipe model with RANSAC, the authors of~\cite{oh2021automated} find candidate points by evaluating local quadric surfaces. An approach from~\cite{lee2013skeleton} uses skeletonisation to find the center axes of cylinders and estimates their parameters as a second step. Neural networks are used as part of the pipeline in~\cite{agapaki2020cloi, youngdoo2020automatic}. The principal difference between the tasks tackled in the aforementioned articles and the task of segmenting a log from a point cloud is that the shape of a log is not as strictly defined as that of a man-made artificial object, requiring a more flexible and complex definition.

\subsection{Point Cloud Processing Using Neural Networks}

% Projection based methods, refs
There exist several approaches to processing point clouds with the help of neural networks. The main difficulty is the lack of structure that is heavily used in, for example, \gls{2D} image processing. A common approach would be to reduce complex problem of working with unstructured \gls{3D} data to a relatively simpler problem of processing \gls{2D} images. This could be achieved by projecting the points, as in~\cite{boulch2018snapnet, milioto2019rangenet, zhang2020polarnet}. Such approach is better suited for the analysis of LiDAR-based scene recognition, where the point cloud is collected from a single source (i.e. a single point in space).

% Volumetric based methods + convolution, refs
Since convolution operation itself can be easily extended to three dimensions, the only obstacle to applying \glspl{CNN} to point clouds is the lack of structure. This can be rectified by voxelising the data into a \gls{3D} grid, making it possible to apply \glspl{CNN}, as proposed in~\cite{maturana2015voxnet}. However, the distribution of points in \gls{3D} space is not uniform, leading to large empty areas, requiring memory and providing no useful information. This can be mitigated by using hierarchical structures, such as octrees~\cite{riegler2017octnet}, allowing to encode more information in dense areas and not spend much resources on empty regions.
Alternatively, the convolution operation itself can be adapted to the unstructured \gls{3D} data case. For example, KPConv~\cite{kpconv} uses linear interpolation to apply the convolution to a small circular area around each point. PAConv~\cite{xu2021paconv} uses a special ScoreNet model to construct convolutional kernels online from the weight bank, which then can be applied to perform convolution, i.e. convolutional kernels are not learned during the training, but constructed during inference. It is also possible to treat the point cloud as a graph, and use graph convolutions, as proposed in~\cite{wang2019dynamic, landrieu2018large}.

% Point-based methods
Instead of trying to transform the unstructured data into a structured representation, it is also possible to design methods that would be invariant to point ordering, allowing to process each point independently. 
The seminal paper~\cite{pointnet} introduced PointNet, a network that uses a small \gls{MLP} to encode each point and then pool the information into a global descriptor.
The main problem is that the local relationship between points is not accounted for, missing crucial information. That is rectified in PointNet++~\cite{pointnetpp}, which divides points into small local neighborhoods, processes and pools them in a hierarchical manner.

The transformer architecture, initially introduced for the natural language processing, have been used to achieve state-of-the-art results in almost all areas of deep learning, point cloud processing being no different. The Point Transformer~\cite{zhao2021point} uses self-attention with position embeddings on local regions around each point by using k nearest neighbours. The transformer layer follows the same general structure for self-attention as in other application-areas, such as 2D image processing. Since the introduction of the original Point Transformer, two updated versions: Point Transformer V2~\cite{wu2022point} and Point Transformer V3~\cite{wu2024point} have been proposed as well. Point Transformer V2~\cite{wu2022point} added grouped vector attention and the partition-based pooling strategy to improve computation speed and better capture spatial information. In addition, the authors have also modified position encoding by adding an \gls{MLP} to calculate a multiplier for the relation vector. Apart from improved performance on a variety of different tasks, Point Transformer V3~\cite{wu2024point} achieves a major improvement in terms of speed and memory consumption, thanks to serialized neighbor mapping. Such approach sacrifices some precision to scale the method to much larger point clouds, increasing the receptive field from 16 to 1024 points.

\section{Proposed Method}

The main objective of this study is to create a method for unsupervised segmentation of point clouds of logs acquired in a sawmill environment. %As a secondary objective, the proposed method is also used to estimate the centreline of a log. 
The method uses the Point Transformer~\cite{zhao2021point} architecture as a starting point. The performance and scalability improvements of V2~\cite{wu2022point} and V3~\cite{wu2024point} are significant, they are mostly relevant for large \gls{3D} scenes, the proposed method is intended to be applied to laser scans of a single object. The loss includes computation of k nearest neighbours, which are already computed in Point Transformer V1, but omitted in subsequent versions, making V1 the natural choice for this method. However, the loss function could as easily be used for training any other model capable of working with point clouds.

For each point $p_i$ from the input point cloud, the network outputs 4 values: a weight $w_i$ and a \gls{3D} centreline vector $\rho_i$. The sigmoid function is applied to the weight values, while the values of centreline vectors are taken without any modification. The centreline vectors are supposed to point from each point to the closest point on the centreline of a log. The main contribution of this paper is the proposal of a special unsupervised loss function that utilises geometrical properties of a log shape. 
The loss function is defined as follows
\begin{equation}
    \mathcal{L} = \lambda_\textrm{fit} \mathcal{L}_\textrm{fit} + \lambda_\rho \mathcal{L}_{\rho} + \lambda_\sigma \mathcal{L}_\sigma + \lambda_\textrm{plane} \mathcal{L}_\textrm{plane} + \lambda_\textrm{n} \mathcal{L}_\textrm{n} + \lambda_W \mathcal{L}_W,
\end{equation}
where $\mathcal{L}_\textrm{fit}$ is the fit loss term, $\mathcal{L}_{\rho}$ is the distance loss term, $\mathcal{L}_\sigma$ is the deviation loss term, $\mathcal{L}_\textrm{plane}$ is plane loss term, $\mathcal{L}_\textrm{n}$ is normal loss term, $\mathcal{L}_W$ is the weights loss term and $\lambda_\textrm{fit}$, $\lambda_\rho$, $\lambda_\sigma$, $\lambda_\textrm{plane}$, $\lambda_\textrm{n}$, $\lambda_W$ are corresponding weights for each loss term. Before describing every loss term in detail, it is important to mention that the weights $w_i$ are used for the calculation of all average values described further as such
\begin{equation}
    \Bar{x} = \frac{\sum_i^n w_i x_i}{\sum_i^n w_i},  
\end{equation}
where $\Bar{x}$ is the weighted average of some values $x_i$. 

\subsection{Fit loss term}
Let us define the closest centreline point for each point $p_i$ as 
\begin{equation}
    c_i = p_i +\rho_i.
\end{equation}
An important assumption is that the point cloud of a log is aligned length-wise along the $x$-axis, meaning that $x$ coordinates of those centreline points can be used to fit a curve (or a line) describing the centreline using polynomial (or linear) regression. A curve might be more desirable depending on the general curvature of logs in question, but even a line can be a good enough approximation for most cases. The curve is found by solving weighted least square regression for both $y$ and $z$ values depending on $x$:
\begin{equation}
    A = (X^T W X)^{-1} X^T W Y,
\end{equation}
where $X$ is a Vandermonde matrix of $x$ values of centreline points $c_i$, $W$ is a diagonal matrix of weights $w_i$ and $Y$ is a $2 \times n$ matrix  containing all $y$ and $z$ coordinates of centreline points on the first and second columns respectively. Then, $x$ coordinates of centreline points and the curve $A$ are used to evaluate new $y$ and $z$ coordinates, which together with $x$ coordinates can be described as fitted curve points $\hat{c}$. Ideally, centreline points $c_i$ should be close to the fitted curve points, therefore the fit loss term is defined as
\begin{equation}
    \mathcal{L}_\textrm{fit} = \frac{\sum_i^n w_i \norm{c_i - \hat{c}_i}}{\sum_i^n w_i}.
\end{equation}
This term is minimised when all centrepoints are lying on a curve.

\subsection{Distance loss term}

Distance term is responsible for ensuring that the output centreline points $c_i$ are as close as possible to their respective points $p_i$. Therefore, the term is simply a weighted mean of all distances to the centreline, i.e. 
\begin{equation}
    \mathcal{L}_\rho = \frac{\sum_i^n w_i \norm{\rho_i}}{\sum_i^n w_i}.
\end{equation}

Distance loss term and fit line term are supposed to act in a tug-of-war fashion to each other, simultaneously moving the centreline points closer together while keeping them close to their respective input points. 

\subsection{Deviation loss term}

An additional constraint can be applied to the distances to the centreline. In a case of a perfectly cylindrical structure, it is reasonable to expect all distances to the centreline to be equal. This could be enforced by including a loss term that would correspond to the deviation of centreline distances $\norm{\rho_i}$ from the mean value. However, wooden logs usually taper closer to the top - a natural property of the growing process. Therefore, a similar condition can be enforced by utilising detrended values of centreline distances. In a similar fashion to the fit loss term, a line can be fitted to $x_i$ values and centreline distance values $\norm{\rho_i}$ to estimate $\widehat{\norm{\rho_i}}$. Deviation loss term is then defined as
\begin{equation}
    \mathcal{L_\sigma} = \frac{\sum_i^n w_i (\norm{\rho_i} - \widehat{\norm{\rho_i}})^2}{\sum_i^n w_i}.
\end{equation}

\subsection{Plane loss term}

The goal is to segment the surface of the log. Therefore, the shape of the local neighbourhood around each surface point should correspond to that of a smooth manifold. By using small enough neighbourhoods, the manifolds can be thought of as planes. Information about the shape of a neighbourhood of point $p_i$ can be extracted by calculating eigenvalues $\lambda_1, \lambda_2, \lambda_3$ of the weighted covariance matrix. When $\lambda_1 \approx \lambda_2 > \lambda_3$, the points are distributed on a plane. Plane loss term for each neighbourhood $Q_i$ is defined to enforce this inequality as
\begin{equation}
   \mathcal{L}_\textrm{plane}(Q_i) = 1 - \frac{\lambda_2}{\lambda_1} + \frac{\lambda_3}{\lambda_2}.
\end{equation}

The final plane loss term is the weighted mean of neighbourhood terms 
\begin{equation}
    \mathcal{L}_\textrm{plane} = \frac{\sum_i^n w_i \mathcal{L}_\textrm{plane}(Q_i)}{\sum_i^n w_i}.
\end{equation}

\subsection{Normal loss term}

It can be assumed that the normal to the log's surface must be collinear to the vector pointing to the centre of the log. If the points in a local neighbourhood lie on a plane, then eigenvector $v_3$ corresponding to the smallest eigenvalue $\lambda_3$ can be used to estimate the normal to that plane. Then, assuming that $v_3$ is a unit vector, the normal loss term for the neighbourhood $Q_i$ is defined as
\begin{equation}
   \mathcal{L}_\textrm{n}(Q_i) = 1 - \abs{\frac{\rho_i \cdot v_3}{\norm{\rho_i}}}.
\end{equation}

In the same vein as plane loss term, the final normal loss term is defined as a weighted mean of normal loss terms for all neighbourhoods
\begin{equation}
    \mathcal{L}_\textrm{n} = \frac{\sum_i^n w_i \mathcal{L}_\textrm{n}(Q_i)}{\sum_i^n w_i}.
\end{equation}

\subsection{Weights loss term}

Finally, in order to avoid a trivial solution $\forall i, w_i = 0$ and ensure that as many inlier points are found as possible, a weights loss term is defined as 
\begin{equation}
    \mathcal{L}_W = 1 - \frac{1}{n}\sum_i^n w_i.
\end{equation}

\section{Experiments}

\subsection{Data}

The data for the experiments were collected from a real sawmill environment and consisted of point clouds of 200 debarked softwood logs. Only Scots pine (\textit{Pinus sylvestris}) logs were used. The point clouds can be divided into 3 separate subsets, each corresponding to a different data collection session from 3 different sawmills. Some logs were scanned several times, producing several point clouds of the same log. More information about the data is presented in \cref{tab:data}. The data has been split into training, validation and test sets using 50\% - 25\% - 25\% split, making sure that all scans of the same log were present only in one subset. Due to the way the data was collected, all point clouds are aligned in such a way that the log is aligned along the $x$ axis. In a more general setting, the same can be achieved by using \gls{PCA} to align the points such that the $x$ axis corresponds to the direction with the most variance, since the logs are, generally, longer than they are wide. 

\begin{table}
\centering
\caption{Data used in experiments. The density is defined as the distance to the closest neighbour in mm.}\label{tab:data}
\begin{tabular}{@{}ccccc@{}}
\toprule
Subset & Logs & \begin{tabular}[c]{@{}c@{}}\# of scans \\ per log\end{tabular} & \begin{tabular}[c]{@{}c@{}}Mean \# \\ of points\end{tabular} & \begin{tabular}[c]{@{}c@{}}Mean \\ density\end{tabular} \\ 
\midrule
A & 1-50                       & 4                                                                       & 591 990                                                                              & 1.366217     \\
B & 51-100                     & 2                                                                       & 700 615                                                                              & 1.6308846    \\
C & 101-200                    & 1                                                                       & 357 187                                                                              & 1.4246505    \\ 
\bottomrule
\end{tabular}

\end{table}

For the purpose of evaluating the network during validation and testing, ground truth segmentation labels were first generated using the method proposed in~\cite{zolotarev2020modelling} and then manually corrected. \gls{IoU} is used as an evaluation metric during the testing along with precision and recall for a more detailed analysis of the method performance.

\subsection{Training}

Point Transformer~\cite{zhao2021point} has been chosen for the network architecture. Point-based approach is preferable, since the voxelisation depends on the choice of resolution and can lead to a loss of finer details and projection-based methods are ill-suited for this type of data, i.e. point clouds of a single object. 

ADAM~\cite{adam} optimiser has been used with a learning rate of $0.001$. Loss weights were chosen by using a Hyperband~\cite{hyperband} hyperparameter optimisation. The best weights are $\lambda_{\textrm{fit}}=0.26, \lambda_\rho=0.12,\lambda_\sigma=0.18, \lambda_{\textrm{plane}}=0.12, \lambda_n=0.09, \lambda_W=0.23$. Additionally, due to the fact that the log is already aligned along the $x$ axis, the resulting centreline vector $\rho$ is only optimised along $y$ and $z$ with $x$ offset permanently set to $0$. The size of the neighbourhood used for the calculation of plane and normal terms is $256$.
% Preprocessing
Since the length and radius of the logs can vary quite a bit, it is important to minimise the bias introduced by it. All point clouds are preprocessed by normalising the coordinates. First, the points are centered using median values to minimise the influence of the noise.
Using the main assumption that the log is aligned lengthwise along the $x$ axis, $y$ and $z$ dimensions are divided by \gls{MAD} of the $\sqrt{y^2 + z^2}$ distances. This step essentially scales the radius of the log to be equal to 1. The $x$ coordinates are scaled by \gls{MAD} of the centered $x$ coordinates, bringing the length of the log to $1$ as well. The $y$ and $z$ coordinates are using the same scale in order to better preserve original shape and aspect ratio of the log.
Additionally, each log point cloud is divided into several batches along the length of the log. As data augmentation, only randomly selected $\frac{2}{3}$ of the points from the batch are used for training. 

\subsection{Ablation study}

With such a high number of different terms in the loss function, it is necessary to estimate the influence of each of them on the final output. Out of the 6 terms, 3 of them are necessary to achieve convergence and avoid trivial solutions: fit term $\lambda_{\textrm{fit}}$, distance  term $\lambda_{\rho}$ and weight term $\lambda_W$. Fit term ensures that the points that estimated vectors $\rho_i$ point to are all lying on a curve, while the distance term minimises the distance from those points to the log points, making sure that the found curve corresponds to the centerline of a log. And most importantly, weight term is necessary to avoid the trivial solution of removing all points and acts as a counterweight to the rest of the terms. Deviation term $\lambda_{\sigma}$, plane term $\lambda_{\textrm{plane}}$ and normal term $\lambda_{\textrm{n}}$ are all optional, so an ablation study has been performed by training the network with the same parameters and dataset split, but using different combinations of those terms. 

\begin{figure*}[tb!]
  \centering
  \begin{subfigure}{0.32\linewidth}
    \includegraphics[width=1\textwidth]{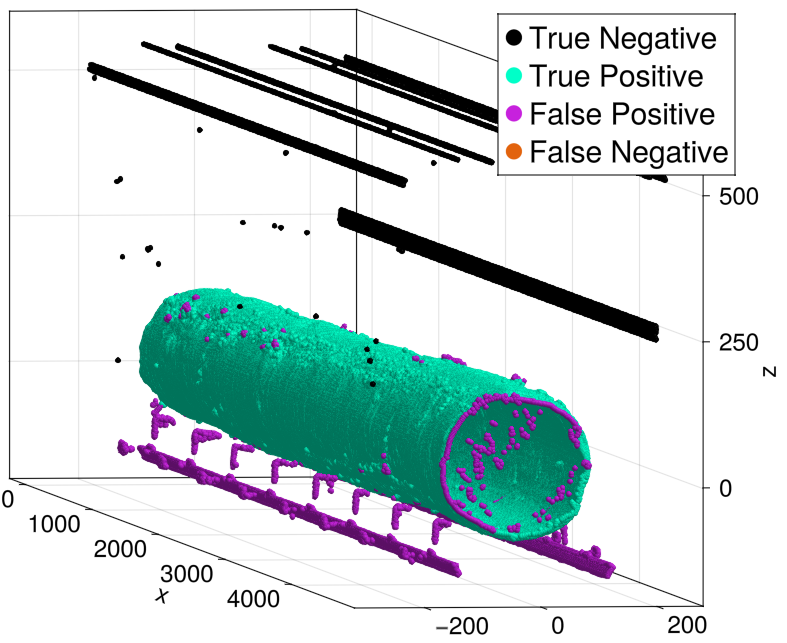}
    \caption{Deviation term used.}
  \end{subfigure}
  % \hfill
  \begin{subfigure}{0.32\linewidth}
    \includegraphics[width=1\textwidth]{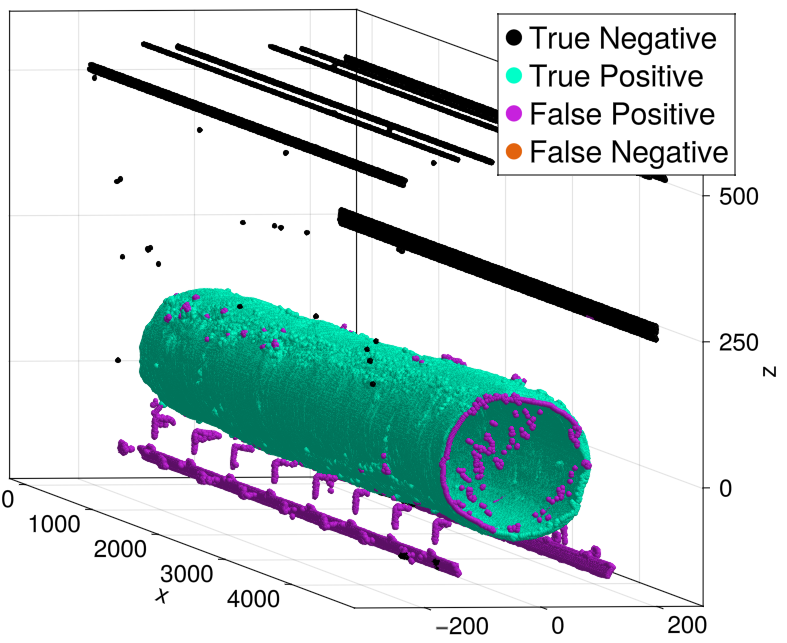}
    \caption{Normal term used.}
  \end{subfigure}
  % \hfill
  \begin{subfigure}{0.32\linewidth}
    \includegraphics[width=1\textwidth]{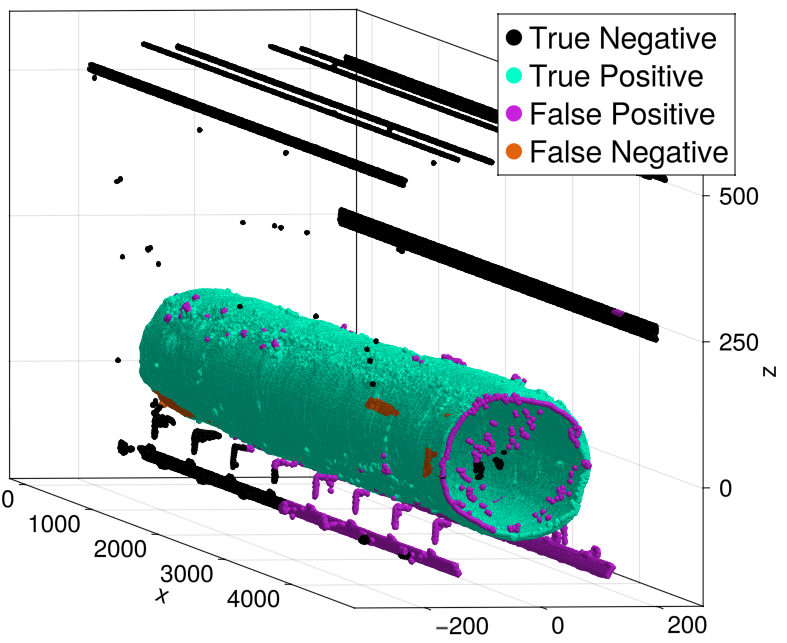}
    \caption{Plane term used.}
  \end{subfigure}
  % \hfill
  \begin{subfigure}{0.32\linewidth}
    \includegraphics[width=1\textwidth]{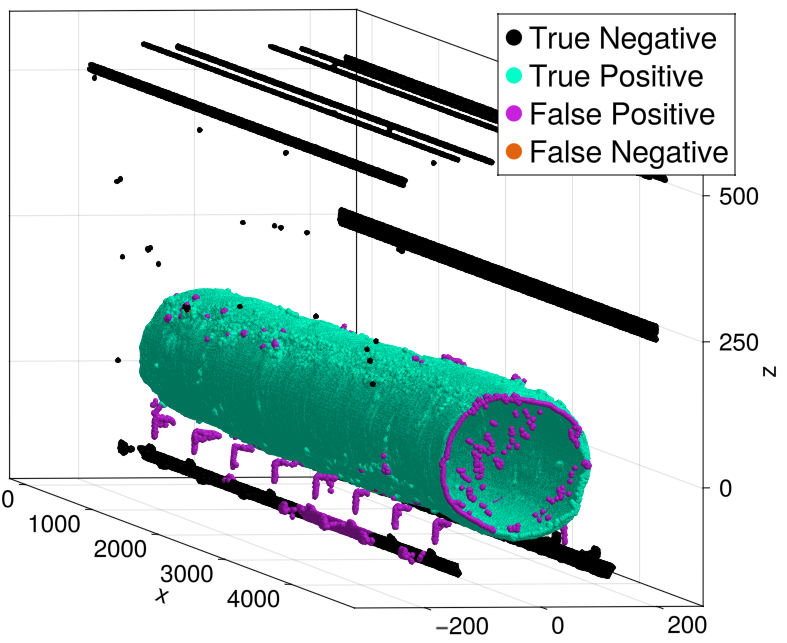}
    \caption{Plane and normal terms used.}
  \end{subfigure}
  % \hfill
  \begin{subfigure}{0.32\linewidth}
    \includegraphics[width=1\textwidth]{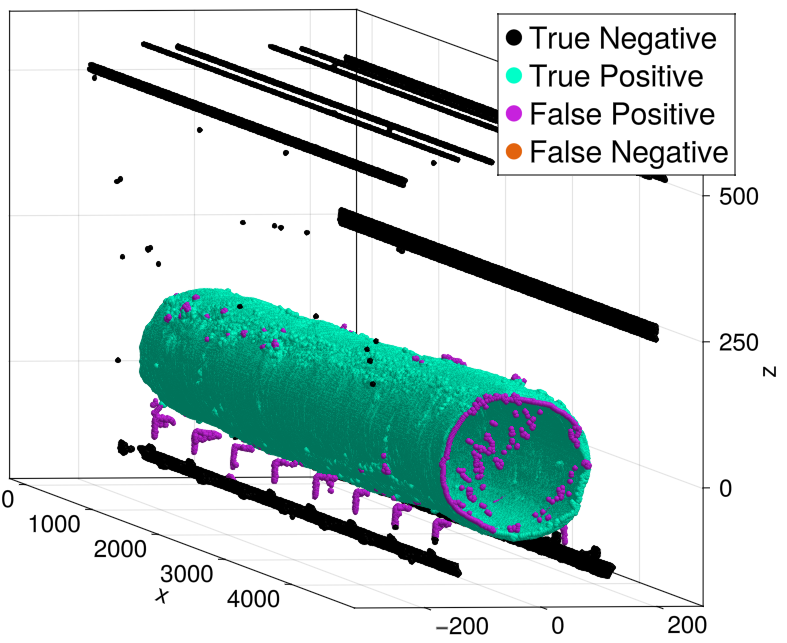}
    \caption{Deviation and normal terms used.}
  \end{subfigure}
  \begin{subfigure}{0.32\linewidth}
    \includegraphics[width=1\textwidth]{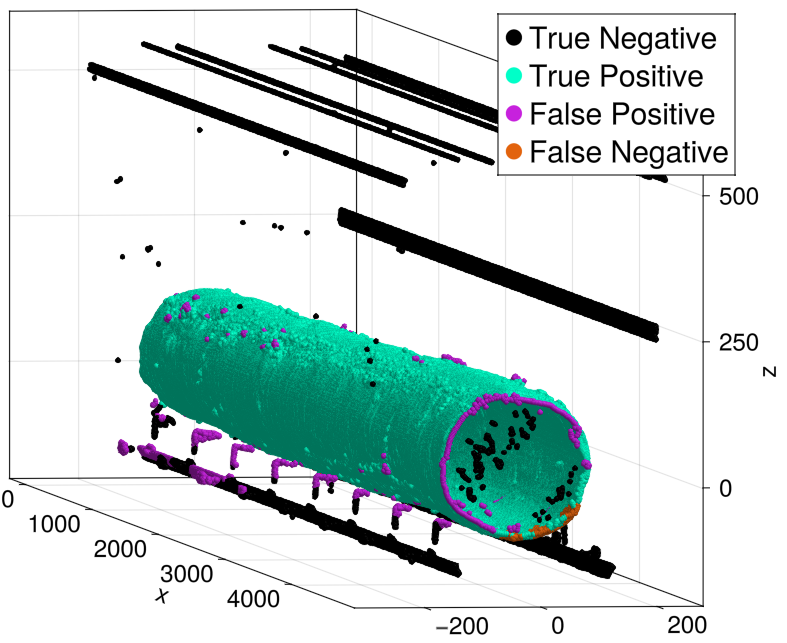}
    \caption{Deviation and plane terms used.}
  \end{subfigure}\\
  % \hfill
  \begin{subfigure}{0.32\linewidth}
    \includegraphics[width=1\textwidth]{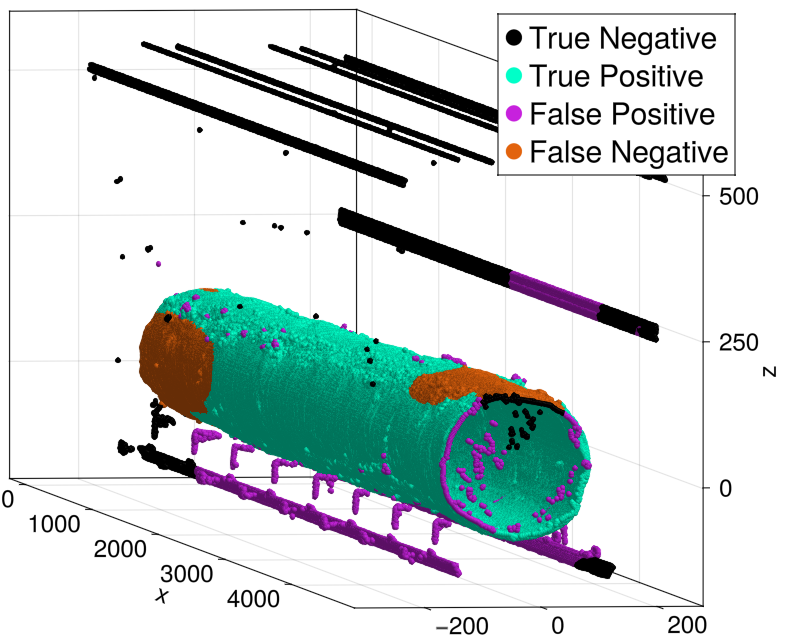}
    \caption{No optional terms.}
  \end{subfigure}
  % \hfill
  \begin{subfigure}{0.32\linewidth}
    \includegraphics[width=1\textwidth]{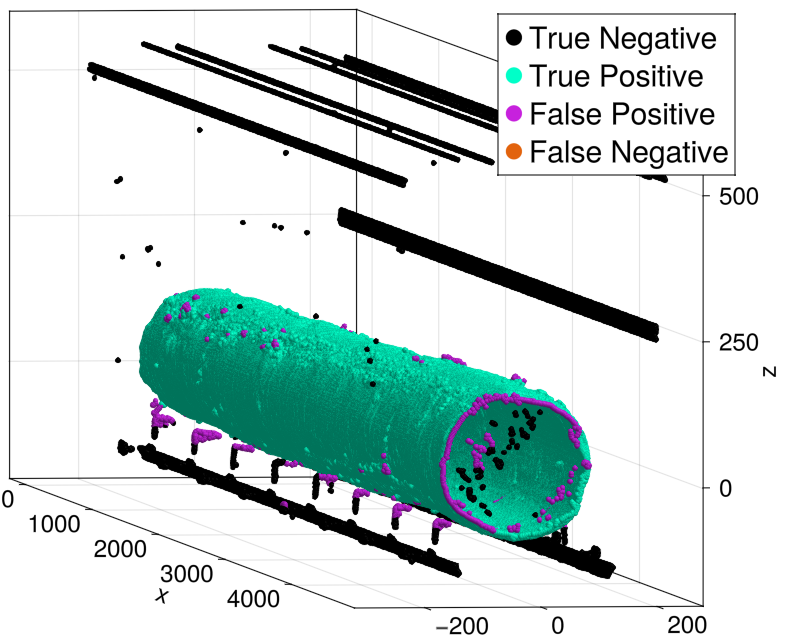}
    \caption{All optional terms used.}
  \end{subfigure}
  \caption{Examples of the method application to the same log while training with various subsets of loss terms.}
  \label{fig:ablation}
\end{figure*}

The quantitative results are presented in \cref{tab:ablation}. The highest \gls{IoU} is achieved by using all optional terms, along with the highest precision. The recall in most cases fluctuates around $98\%$, with the exception of the case where all optional terms are omitted. Such high recall 
is the result of the inherent class imbalance, i.e. the number of outliers relative to the log points is quite low. The qualitative results are presented on \cref{fig:ablation}.

\begin{table}
    \caption{Mean results on the test set for different combinations of optional loss terms. The best results for each column are highlighted in bold.}
    \label{tab:ablation}
    \centering
\begin{tabular}{@{}cccccc@{}}
  \toprule
     Deviation & Plane & Normal & Precision & Recall & \acrshort{IoU} \\
  \midrule
    \xmark&\xmark&\xmark& 89.73\% & 91.36\% & 82.64\% \\
    \xmark&\xmark&\cmark& 92.63\% & \textbf{98.52}\% & 91.35\% \\
    \xmark&\cmark&\xmark& 93.37\% & 97.61\% & 91.26\% \\
    \xmark&\cmark&\cmark& 97.45\% & 98.35\% & 95.87\% \\
    \cmark&\xmark&\xmark& 91.67\% & 98.51\% & 90.36\% \\
    \cmark&\xmark&\cmark& 98.27\% & 98.51\% & 96.83\% \\
    \cmark&\cmark&\xmark& 98.24\% & 97.75\% & 96.07\% \\
    \cmark&\cmark&\cmark& \textbf{98.88}\% & 98.45\% & \textbf{97.35}\% \\
  \bottomrule
\end{tabular}
\end{table}

\subsection{Results}

\begin{figure*}[tb!]
  \centering
  \begin{subfigure}{0.32\linewidth}
    \includegraphics[width=1\textwidth]{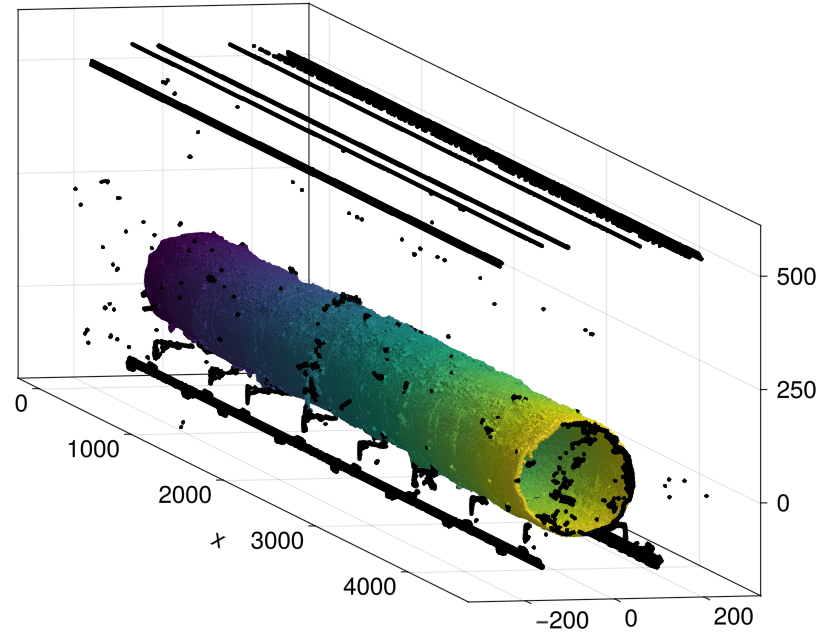}
    \caption{Ground truth for a point cloud from Subset A.}
  \end{subfigure}
  \hfill
  \begin{subfigure}{0.32\linewidth}
    \includegraphics[width=1\textwidth]{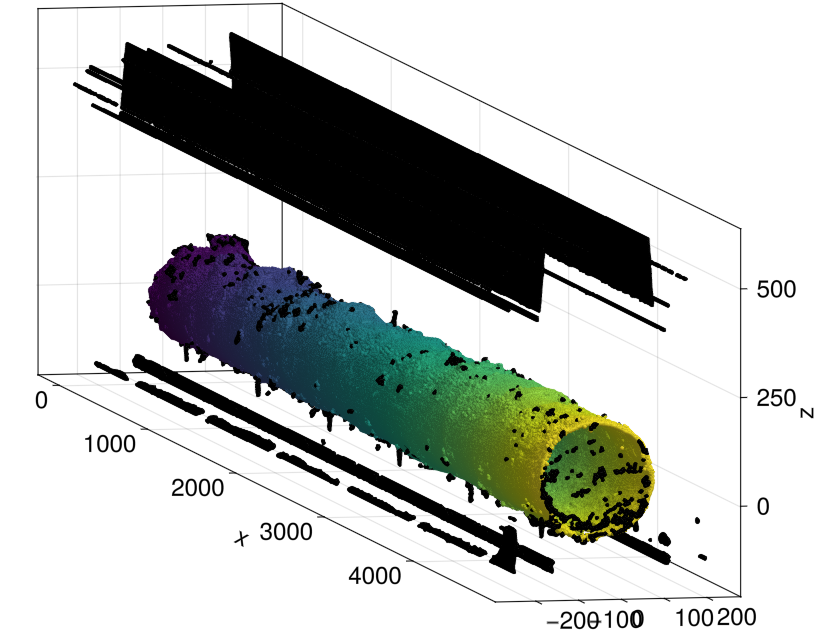}
    \caption{Ground truth for a point cloud from Subset B.}
  \end{subfigure}
  \hfill
  \begin{subfigure}{0.32\linewidth}
    \includegraphics[width=1\textwidth]{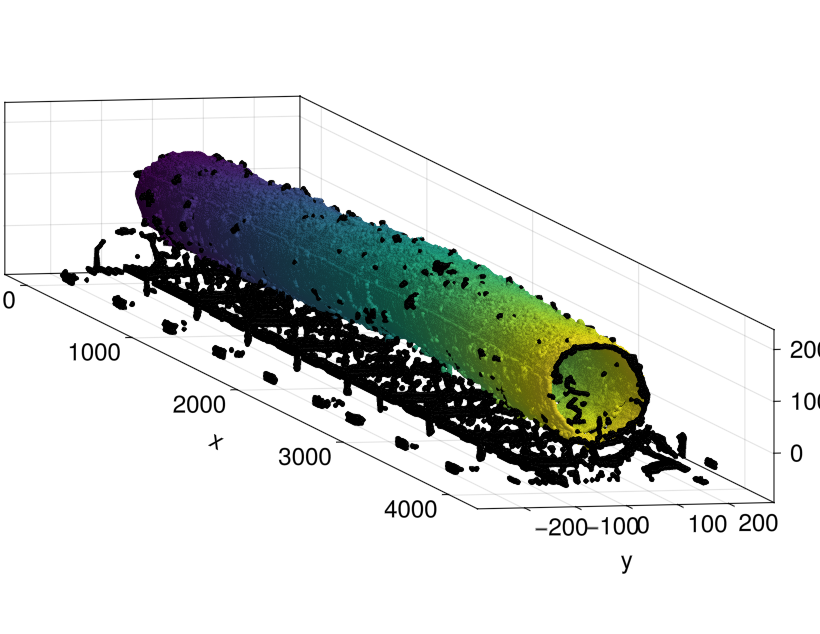}
    \caption{Ground truth for a point cloud from Subset C.}
  \end{subfigure}\\
  \begin{subfigure}{0.32\linewidth}
    \includegraphics[width=1\textwidth]{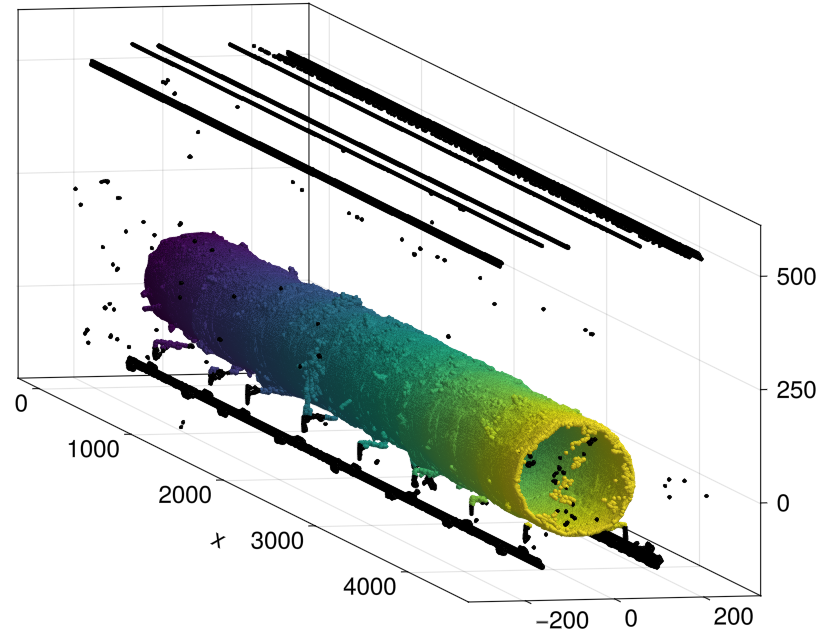}
    \caption{Method output for a point cloud from Subset A.}
  \end{subfigure}
  \hfill
  \begin{subfigure}{0.32\linewidth}
    \includegraphics[width=1\textwidth]{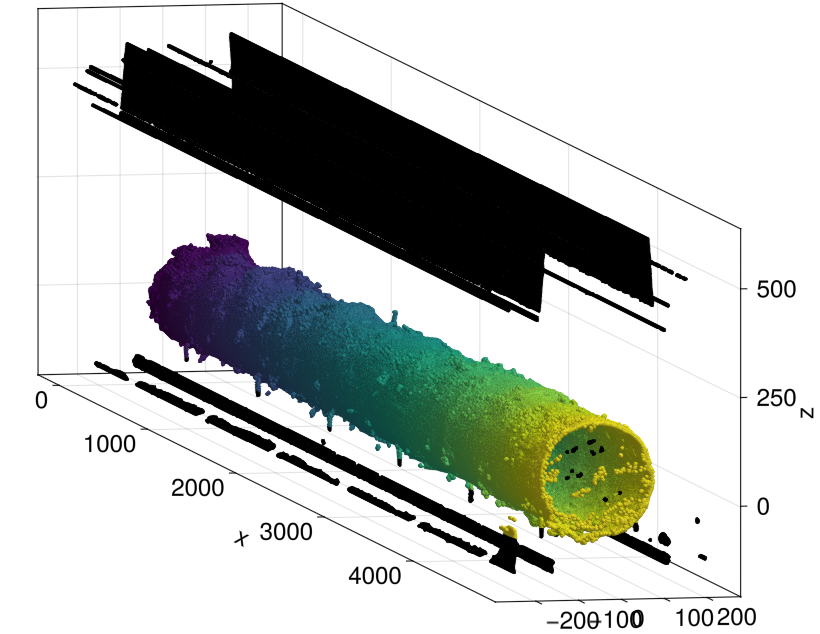}
    \caption{Method output for a point cloud from Subset B.}
  \end{subfigure}
  \hfill
  \begin{subfigure}{0.32\linewidth}
    \includegraphics[width=1\textwidth]{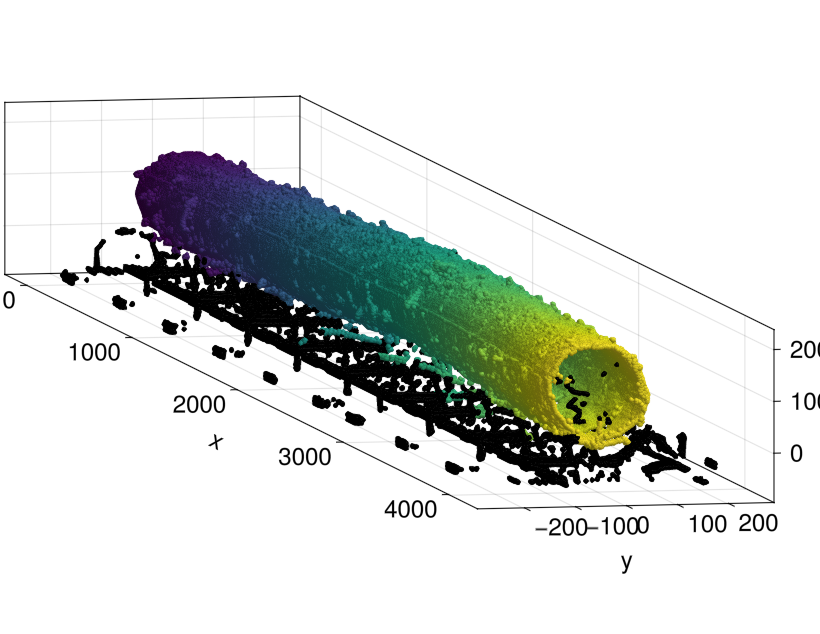}
    \caption{Method output for a point cloud from Subset C.}
  \end{subfigure}\\
  \begin{subfigure}{0.32\linewidth}
    \includegraphics[width=1\textwidth]{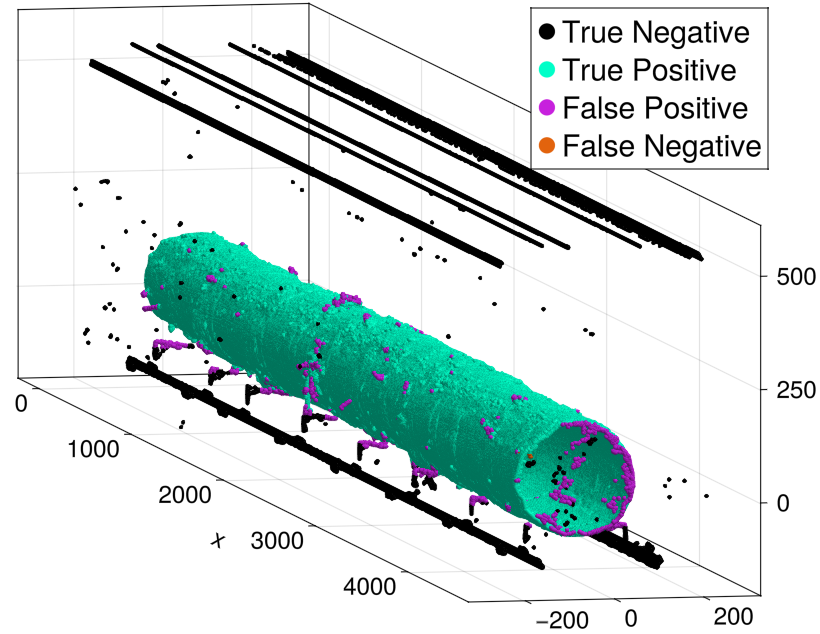}
    \caption{Comparison for a point cloud from Subset A.}
  \end{subfigure}
  \hfill
  \begin{subfigure}{0.32\linewidth}
    \includegraphics[width=1\textwidth]{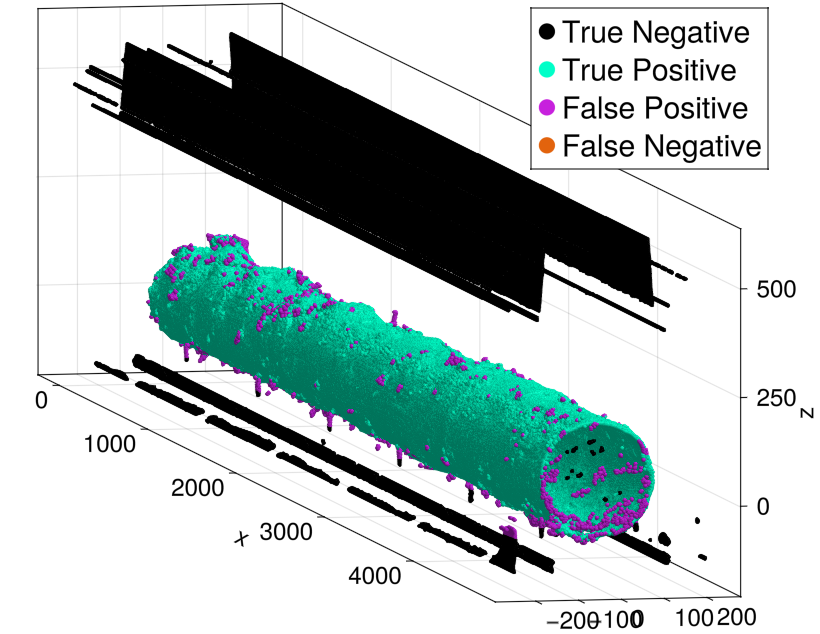}
    \caption{Comparison for a point cloud from Subset B.}
  \end{subfigure}
  \hfill
  \begin{subfigure}{0.32\linewidth}
    \includegraphics[width=1\textwidth]{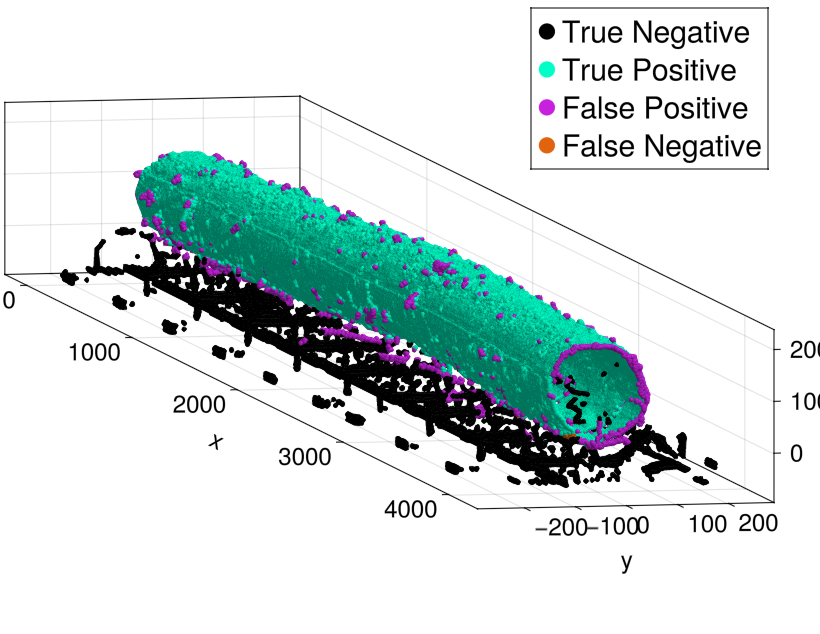}
    \caption{Comparison for a point cloud from Subset C.}
  \end{subfigure}
  \caption{Examples of the output on data from the test set.}
  \label{fig:results}
\end{figure*}

The results of comparing the segmentation performance on the test set are presented in \cref{tab:results}. There is a small variation in the results for each subset due to their heterogeneous nature, with the worst \gls{IoU} of $95.32\%$ for Subset B, while the results for other subsets are more consistent with the \gls{IoU} of around $97\%$. Recall and precision are both quite high, with a minimum score of $96.61\%$.

\begin{table}
    \caption{Mean results on the test set. The results are presented separately for point clouds from each of the subsets and for the whole test set.}
    \label{tab:results}
    \centering
\begin{tabular}{@{}cccc@{}}
  \toprule
   Subset & Precision & Recall & \acrshort{IoU} \\
   \midrule
  A & 99.44\% & 98.52\% & 97.97\% \\
  B & 98.74\% & 96.81\% & 95.60\% \\
  C & 97.97\% & 99.88\% & 97.86\% \\
  \midrule
  Whole test set & 98.88\% & 98.45\% & 97.35\% \\
  \bottomrule
\end{tabular}

\end{table}

Examples of method application to the point clouds from the test set are presented in \cref{fig:results}. Due to the data being collected in 3 separate sessions, the distribution of the outliers varies significantly between the subsets of data. Consistent with the high recall values, the whole log is segmented for all presented samples. The problematic cases are connected to the metal railing carrying the log and noisy points that are very close to the log surface. Metal railings appear only on one side of the point cloud and have a clear periodic structure, and therefore, they can be eliminated with post-processing. Noisy points close to log surface are challenging to distinguish from natural parts of the log such as bark. Therefore, part of the incorrectly segmented points can actually be mislabeled in the ground truth.
%However, the method fails to remove parts of the metal railing and noisy points that are too close to the log. In the case of a log from Subset B, even relatively faraway railings were erroneously segmented.

% \section{Acknowledgements}
% The research was supported by the Finnish Research Impact Foundation (project number 241).

% \input{sec/1_intro}
% \input{sec/2_formatting}
% \input{sec/3_finalcopy}
% \input{sec/X_suppl}

\section{Conclusion}

The main goal of this study was to create a fully unsupervised end-to-end method for the segmentation of logs from point clouds. The focus on being fully unsupervised greatly enhances the applicability of the proposed approach to the real-world industrial setting, where collecting and annotating large amounts of data requires too many resources to be feasible. The method utilises Point Transformer neural network to process point clouds. The network is trained using a carefully designed loss function that utilises geometrical properties commonly found in wooden logs to separate log points from outliers. Additionally, the network generates centreline points as a byproduct, which could be utilised for further processing of the log. One example would be the generation of a surface heightmap, which is a widely used way of detecting defects on the surface of a log. The results indicate that while the method is able to correctly segment most log points, it also struggles to remove outliers close to the actual log surface. The method could potentially benefit from increasing the size of a training dataset. Additional loss terms could be added to specialise on hard areas, \eg separating support railings touching the surface of the log from the actual log points.

% ---- Bibliography ----
%
% BibTeX users should specify bibliography style 'splncs04'.
% References will then be sorted and formatted in the correct style.
%
\bibliographystyle{splncs04}
\bibliography{short}

\begin{thebibliography}{10}
\providecommand{\url}[1]{\texttt{#1}}
\providecommand{\urlprefix}{URL }
\providecommand{\doi}[1]{https://doi.org/#1}

\bibitem{agapaki2020cloi}
Agapaki, E., Brilakis, I.: Cloi-net: Class segmentation of industrial facilities’ point cloud datasets. Advanced Engineering Informatics  \textbf{45},  101121 (2020). \doi{https://doi.org/10.1016/j.aei.2020.101121}, \url{https://www.sciencedirect.com/science/article/pii/S1474034620300902}

\bibitem{batrakhanov2021virtual}
Batrakhanov, D., et~al.: Virtual sawing using generative adversarial networks. In: International Conference on Image and Vision Computing New Zealand (2021). \doi{10.1109/IVCNZ54163.2021.9653436}

\bibitem{boulch2018snapnet}
Boulch, A., et~al.: Snapnet: 3d point cloud semantic labeling with 2d deep segmentation networks. Computers \& Graphics  \textbf{71},  189--198 (2018)

\bibitem{ester1996density}
Ester, M., et~al.: {A Density-Based Algorithm for Discovering Clusters in Large Spatial Databases with Noise}. In: {KDD} ({1996})

\bibitem{young2019fast}
Jin, Y.H., Lee, W.H.: Fast cylinder shape matching using random sample consensus in large scale point cloud. Applied Sciences  \textbf{9}(5) (2019). \doi{10.3390/app9050974}, \url{https://www.mdpi.com/2076-3417/9/5/974}

\bibitem{youngdoo2020automatic}
Kim, Y., et~al.: Automatic pipe and elbow recognition from three-dimensional point cloud model of industrial plant piping system using convolutional neural network-based primitive classification. Automation in Construction  \textbf{116},  103236 (2020). \doi{https://doi.org/10.1016/j.autcon.2020.103236}, \url{https://www.sciencedirect.com/science/article/pii/S092658051931547X}

\bibitem{adam}
Kingma, D.P., Ba, J.: Adam: A method for stochastic optimization. In: 3rd International Conference on Learning Representations (2015)

\bibitem{kretschmer2013new}
Kretschmer, U., et~al.: {A new approach to assessing tree stem quality characteristics using terrestrial laser scans}. {Silva Fennica}  \textbf{{47}} ({2013}). \doi{10.14214/sf.1071}

\bibitem{landrieu2018large}
Landrieu, L., Simonovsky, M.: Large-scale point cloud semantic segmentation with superpoint graphs. In: IEEE Conference on Computer Vision and Pattern Recognition (June 2018)

\bibitem{lee2013skeleton}
Lee, J., et~al.: Skeleton-based 3d reconstruction of as-built pipelines from laser-scan data. Automation in Construction  \textbf{35},  199--207 (2013). \doi{https://doi.org/10.1016/j.autcon.2013.05.009}, \url{https://www.sciencedirect.com/science/article/pii/S0926580513000769}

\bibitem{hyperband}
Li, L., et~al.: Hyperband: A novel bandit-based approach to hyperparameter optimization. The Journal of Machine Learning Research  \textbf{18},  185:1--185:52 (2017), \url{http://jmlr.org/papers/v18/16-558.html}

\bibitem{maturana2015voxnet}
Maturana, D., Scherer, S.: Voxnet: A 3d convolutional neural network for real-time object recognition. In: International Conference on Intelligent Robots and Systems. pp. 922--928 (2015). \doi{10.1109/IROS.2015.7353481}

\bibitem{milioto2019rangenet}
Milioto, A., et~al.: Rangenet++: Fast and accurate lidar semantic segmentation. In: IEEE/RSJ International Conference on Intelligent Robots and Systems. pp. 4213--4220 (2019)

\bibitem{oh2021automated}
Oh, I., Ko, K.H.: Automated recognition of 3d pipelines from point clouds. The Visual Computer  \textbf{37}(6),  1385–1400 (2021). \doi{10.1007/s00371-020-01872-y}, \url{https://doi.org/10.1007/s00371-020-01872-y}

\bibitem{patil2017adaptive}
Patil, A.K., et~al.: An adaptive approach for the reconstruction and modeling of as-built 3d pipelines from point clouds. Automation in Construction  \textbf{75},  65--78 (2017)

\bibitem{pfeifer2004automatic}
Pfeifer, N., et~al.: {Automatic Reconstruction of Single Trees from Terrestrial Laser Scanner Data}. {International Archives of Photogrammetry, Remote Sensing and Spatial Information Sciences}  \textbf{{35}} ({2004})

\bibitem{pointnetpp}
Qi, C.R., et~al.: Pointnet++: Deep hierarchical feature learning on point sets in a metric space. In: Advances in Neural Information Processing Systems. vol.~30 (2017), \url{https://proceedings.neurips.cc/paper/2017/file/d8bf84be3800d12f74d8b05e9b89836f-Paper.pdf}

\bibitem{pointnet}
Qi, C.R., et~al.: Pointnet: Deep learning on point sets for 3d classification and segmentation. In: IEEE Conference on Computer Vision and Pattern Recognition (July 2017)

\bibitem{rais2017use}
Rais, A., et~al.: The use of the first industrial x-ray ct scanner increases the lumber recovery value: case study on visually strength-graded douglas-fir timber. Annals of Forest Science  \textbf{74}(2), ~28 (2017). \doi{10.1007/s13595-017-0630-5}, \url{https://doi.org/10.1007/s13595-017-0630-5}

\bibitem{riegler2017octnet}
Riegler, G., Ulusoy, O.A., Geiger, A.: Octnet: Learning deep 3d representations at high resolutions. In: IEEE Conference on Computer Vision and Pattern Recognition (July 2017)

\bibitem{romanengo2023recognizing}
Romanengo, C., Raffo, A., Biasotti, S., Falcidieno, B.: Recognizing geometric primitives in 3d point clouds of mechanical cad objects. Computer-Aided Design  \textbf{157},  103479 (2023). \doi{https://doi.org/10.1016/j.cad.2023.103479}, \url{https://www.sciencedirect.com/science/article/pii/S0010448523000118}

\bibitem{stangle2015potentially}
St{\"a}ngle, S.M., et~al.: {Potentially increased sawmill yield from hardwoods using X-ray computed tomography for knot detection}. Annals of Forest Science  \textbf{72}(1),  57--65 (2015)

\bibitem{kpconv}
Thomas, H., et~al.: Kpconv: Flexible and deformable convolution for point clouds. In: IEEE/CVF International Conference on Computer Vision. pp. 6410--6419 (2019). \doi{10.1109/ICCV.2019.00651}

\bibitem{thomas2007robust}
Thomas, L., Mili, L.: {A robust GM-estimator for the automated detection of external defects on barked hardwood logs and stems}. Transactions on Signal Processing  \textbf{{55}} ({2007}). \doi{10.1109/TSP.2007.894262}

\bibitem{thomas2006automated}
Thomas, L., et~al.: {Automated detection of severe surface defects on barked hardwood logs}. {Forest Products Journal}  \textbf{{57}} (2006)

\bibitem{thomas2009modeling}
Thomas, R.E.: {Modeling the relationships among internal defect features and external Appalachian hardwood log defect indicators}. {Silva Fennica}  \textbf{{43}} (2009). \doi{10.14214/sf.199}

\bibitem{thomas2013raysaw}
Thomas, R.E.: {RAYSAW: A log sawing simulator for 3D laser-scanned hardwood logs}. In: {Central Hardwood Forest Conference} (2013)

\bibitem{wang2019dynamic}
Wang, Y., et~al.: Dynamic graph cnn for learning on point clouds. ACM Transactions on Graphics  \textbf{38}(5) (2019). \doi{10.1145/3326362}, \url{https://doi.org/10.1145/3326362}

\bibitem{wu2024point}
Wu, X., Jiang, L., Wang, P.S., Liu, Z., Liu, X., Qiao, Y., Ouyang, W., He, T., Zhao, H.: Point transformer v3: Simpler faster stronger. In: IEEE/CVF International Conference on Computer Vision. pp. 4840--4851 (2024)

\bibitem{wu2022point}
Wu, X., Lao, Y., Jiang, L., Liu, X., Zhao, H.: Point transformer v2: Grouped vector attention and partition-based pooling. In: Advances in Neural Information Processing Systems (2022), \url{https://openreview.net/forum?id=I3mLa12s_H}

\bibitem{xu2021paconv}
Xu, M., Ding, R., Zhao, H., Qi, X.: Paconv: Position adaptive convolution with dynamic kernel assembling on point clouds. In: IEEE/CVF Conference on Computer Vision and Pattern Recognition. pp. 3173--3182 (June 2021)

\bibitem{zhang2020polarnet}
Zhang, Y., et~al.: Polarnet: An improved grid representation for online lidar point clouds semantic segmentation. In: IEEE/CVF Conference on Computer Vision and Pattern Recognition. pp. 9601--9610 (2020)

\bibitem{zhao2021point}
Zhao, H., et~al.: Point transformer. In: IEEE/CVF International Conference on Computer Vision. pp. 16259--16268 (2021)

\bibitem{zolotarev2019timber}
Zolotarev, F., et~al.: Timber tracing with multimodal encoder-decoder networks. In: International Conference on Computer Analysis of Images and Patterns. pp. 342--353 (2019)

\bibitem{zolotarev2020modelling}
Zolotarev, F., et~al.: Modelling internal knot distribution using external log features. Computers and Electronics in Agriculture  \textbf{179},  105795 (2020)

\end{thebibliography}
\end{document}